\documentclass[a4paper,10pt]{article}

\usepackage{algorithm}
\usepackage{algorithmic}
\usepackage{subfigure}
\usepackage{boxedminipage}
\usepackage{colortbl}
\usepackage{longtable}
\usepackage{lscape}
\usepackage{amsmath}
\usepackage{amsfonts}
\usepackage{amssymb}
\usepackage{url}

\usepackage[pdftex]{graphicx}
\DeclareGraphicsExtensions{.png,.pdf}

\newcommand{\RR}{\mathbb{R}}

\newcommand{\ZZ}{\mathbb{Z}}

\newcommand{\argmin}{\operatornamewithlimits{argmin}}

\newcommand{\bc}{\boldsymbol{c}}
\newcommand{\bd}{\boldsymbol{d}}
\newcommand{\be}{\boldsymbol{e}}

\newcommand{\bu}{\boldsymbol{u}}

\newcommand{\bx}{\boldsymbol{x}}

\newcommand{\bC}{\boldsymbol{C}}
\newcommand{\bD}{\boldsymbol{D}}

\newcommand{\bI}{\boldsymbol{I}}

\newcommand{\bS}{\boldsymbol{S}}

\newcommand{\bU}{\boldsymbol{U}}

\newcommand{\bX}{\boldsymbol{X}}

\title{PADDLE: Proximal Algorithm for Dual Dictionaries LEarning}

\author{
Curzio Basso, Matteo Santoro, Alessandro Verri\\
Dipartimento di Informatica e Scienze dell'Informazione\\
Universit\`a degli Studi di Genova\\
Genova, 16128 Italy \\
\texttt{\{curzio.basso,santoro,verri\}@disi.unige.it} \\
~\\
Silvia Villa\\
Dipartimento di Matematica\\
Universit\`a degli Studi di Genova\\
Genova, 16128 Italy \\
\texttt{\{silvia.villa\}@dima.unige.it}
}

\date{\today}

\begin{document}
\parindent 0pt

 {{\LARGE \font\disifont=cmsy10 scaled \magstep 3
\newbox\Ball
\setbox\Ball=\hbox{\disifont\char"0F}
\newdimen\BB \BB=\wd\Ball
\advance\BB by -.22\BB
\newbox\dSB
\setbox\dSB=\hbox{\vrule height.5\BB width.5\BB}
\newbox\dVB
\setbox\dVB=\hbox{\vrule height\BB width.2\BB}
\newbox\dHB
\setbox\dHB=\hbox{\vrule height.2\BB width\BB}
\newbox\dWB
\setbox\dWB=\hbox to .02\BB{\hss\vrule height1.01\BB width.04\BB\hss}

\setbox\Ball=\hbox to 0pt{\hss\vbox to
0pt{\vss\kern.135\BB\hbox{\disifont\char"0F}\vss}\hss}

\def\TL{\hbox to 0pt{\hss\vbox to 0pt{\vss\copy\dSB}}}
\def\TR{\hbox to 0pt{\vbox to 0pt{\vss\copy\dSB}\hss}}
\def\BL{\hbox to 0pt{\hss\vbox to 0pt{\copy\dSB\vss}}}
\def\BR{\hbox to 0pt{\vbox to 0pt{\copy\dSB\vss}\hss}}
\def\SquareBrick#1#2#3#4{\vbox to \BB{\vss\hbox to \BB{\hss #1#2#3#4\hss}\vss}}

\def\boxit#1{\vbox{\hrule\hbox{\vrule\vbox{#1}\vrule}\hrule}}
\def\WB#1{\raise#1\BB\copy\dWB}

\def\DISIwave{\boxit{\hbox to 3.4\BB
      {\SquareBrick\TL\TR\BL\BR
       \WB{.005}\WB{.01}\WB{.02}\WB{.04}\WB{.07}\WB{.11}\WB{.14}\WB{.20}%
       \WB{.27}\WB{.35}\WB{.45}\WB{.55}\WB{.65}\WB{.75}\WB{.83}\WB{.90}\WB{.96}%
       \WB{1.01}\WB{1.06}\WB{1.10}\WB{1.13}\WB{1.15}\WB{1.16}\WB{1.17}\WB{1.18}%
       \WB{1.18}\WB{1.17}\WB{1.14}\WB{1.10}\WB{1.03}\WB{.91}\WB{.75}\WB{.50}%
       \WB{.25}\WB{0}\WB{-.25}\WB{-.50}\WB{-.75}\WB{-.91}\WB{-1.03}\WB{-1.10}%
       \WB{-1.14}\WB{-1.17}\WB{-1.18}\WB{-1.19}\WB{-1.20}\WB{-1.19}\WB{-1.18}%
       \WB{-1.16}\WB{-1.15}\WB{-1.13}\WB{-1.10}\WB{-1.06}\WB{-1.01}\WB{-.96}%
       \WB{-.90}\WB{-.83}\WB{-.75}\WB{-.65}\WB{-.55}\WB{-.45}\WB{-.35}\WB{-.27}%
       \WB{-.18}\WB{-.12}\WB{-.08}\WB{-.05}\WB{-.04}\WB{-.03}\WB{-.025}%
       \SquareBrick\TL\TR\BL\BR\hss}}}

\def\DISIlogo{\hbox to 13.8\BB{\baselineskip=0pt
\DISIwave\hfill
\hbox{
 \vbox to 3.4\BB{%
 \hbox{\SquareBrick\TL\TR\BL\BR
        \copy\dVB
         \SquareBrick\TL\TR\BL\BR
          \copy\dVB
           \SquareBrick\TL\BR\BL{\copy\Ball}}\vss
 \hbox{\copy\dHB
        \kern1.4\BB
         \copy\dHB}\vss
 \hbox{\SquareBrick\TL\TR\BL\BR
        \kern1.4\BB
         \SquareBrick\TL\TR\BL\BR}\vss
 \hbox{\copy\dHB
        \kern1.4\BB
         \copy\dHB}\vss
 \hbox{\SquareBrick\TL\TR\BL\BR
        \copy\dVB
         \SquareBrick\TL\TR\BL\BR
          \copy\dVB
           \SquareBrick\TL\TR\BL{\copy\Ball}}%
}}\kern .2\BB
\hbox{
 \vbox to 3.4\BB{%
 \hbox{\SquareBrick\TL\TR\BL\BR}\vss
 \hbox{\copy\dHB}\vss
 \hbox{\SquareBrick\TL\TR\BL\BR}\vss
 \hbox{\copy\dHB}\vss
 \hbox{\SquareBrick\TL\TR\BR{\copy\Ball}}%
}}%
\hbox{
 \vbox to 3.4\BB{\vss
 \hbox{\copy\dVB}%
}}%
\hbox{
 \vbox to 3.4\BB{%
 \hbox{\SquareBrick\TR\BL\BR{\copy\Ball}%
        \copy\dVB
         \SquareBrick\TL\TR\BL\BR
          \copy\dVB
           \SquareBrick\TL\BR\BL\TR}\vss
 \hbox{\copy\dHB}\vss
 \hbox{\SquareBrick\TL\TR\BR{\copy\Ball}%
        \copy\dVB
         \SquareBrick\TL\TR\BL\BR
          \copy\dVB
           \SquareBrick\TL\BL\BR{\copy\Ball}}\vss
 \hbox{\kern 2.4\BB
        \copy\dHB}\vss
 \hbox{\SquareBrick\TL\TR\BR\BL
        \copy\dVB
         \SquareBrick\TL\TR\BL\BR
          \copy\dVB
           \SquareBrick\TL\TR\BL{\copy\Ball}}%
}}\kern .2\BB
\hbox{
 \vbox to 3.4\BB{%
 \hbox{\SquareBrick\TL\TR\BL\BR}\vss
 \hbox{\copy\dHB}\vss
 \hbox{\SquareBrick\TL\TR\BL\BR}\vss
 \hbox{\copy\dHB}\vss
 \hbox{\SquareBrick\TL\TR\BR\BL}%
}}%
}}

  \begin{tabular}[b]{l}
  {\sc Dipartimento di Informatica e}\\[-8pt]
  {\sc Scienze dell'Informazione}\\
  \end{tabular} \hfill \DISIlogo}

  \hrulefill

  \vfill

  \begin{center}
    {\Large \bf PADDLE: Proximal Algorithm for Dual Dictionaries LEarning}
  \ \\
  \ \\
  {\large Curzio Basso, Matteo Santoro, Alessandro Verri, Silvia Villa}
  \end{center}
  \vfill

  {\large
  {\bf Technical Report} \hfill {\bf DISI-TR-2010-06}

  \hrulefill

  DISI, Universit\`a di Genova}

  {\normalsize
  v. Dodecaneso 35,
  16146 Genova, Italy
  \hfill {\tt http://www.disi.unige.it/}}}%
 \vfil\null

\maketitle

\begin{abstract}
Recently, considerable research efforts have been devoted to the design of 
methods to learn from data overcomplete dictionaries for sparse coding.
However, learned dictionaries require the solution of an optimization 
problem for coding new data.
In order to overcome this drawback, we propose an algorithm aimed at 
learning both a dictionary and its {\em dual}: 
a linear mapping directly performing the coding.
By leveraging on proximal methods, our algorithm jointly minimizes the 
reconstruction error of the dictionary and the coding error of its dual; the 
sparsity of the representation is 
induced by an $\ell_1$-based penalty on its coefficients.
The results obtained on synthetic data and real images show that the
 algorithm is capable of recovering the expected dictionaries.
Furthermore, on a benchmark dataset, we show that the image features
obtained from the dual matrix yield state-of-the-art classification
performance while being much less computational intensive.
\end{abstract}

\tableofcontents

\section{Introduction}

The representation of a signal as the superposition of elementary
signals, or {\em atoms}, is the pillar of a number of research fields
and analysis techniques. 
The best-known example of such methods is the Fourier transform, where
the atoms form an orthonormal basis and every signal has a unique
representation. 
Although an orthonormal basis would seem the most natural choice for 
decomposing a signal, overcomplete dictionaries (or {\em frames}) are
nowadays commonplace and their use is both theoretically justified and
supported by experimentally successful applications \cite{mallat09}. 
{\em Tight frames} are a class of overcomplete dictionaries with the
particular property of ensuring that the optimal representation can still 
be recovered, as with orthonormal bases, by means of inner products of the 
signal and the dictionary.

The goal of this paper is to introduce an algorithm -- that we called
PADDLE -- capable of learning from data a dictionary endowed with
properties similar to that of tight frames. 
Indeed, the proposed method generates both the optimal dictionary and its 
(approximate) {\em dual}: a linear operator that decomposes new signals to 
their optimal sparse representations, without the need for solving any
further optimization problem.
We stress that the term dual has been adopted in keeping with the 
terminology used for frames, and does not refer to the concept of duality 
common in optimization.

Over the years considerable effort has been devoted to the design of 
methods for learning optimal dictionaries from data.
Although not yielding overcomplete dictionaries, Principal Component
Analysis (PCA) and its derivatives are at the root of such approaches,
based on the minimization of the error in reconstructing the training data 
as a linear combination of the basis elements.
The seminal work of Olshausen and Field \cite{olshausen97} was the first
to propose an algorithm for learning an overcomplete dictionary
in the field of natural image analysis.
Probabilistic assumptions on the data led to a cost function made up of a
reconstruction error and a sparse prior on the coefficients,
and the minimization was performed by alternating optimizations with 
respect to the coefficients and to the dictionary.
Most subsequent methods are based on this alternating scheme of optimization,
with the main differences being the specific techniques used to induce a 
sparse representation.
Recent advances in compressed sensing and feature selection led to use
$\ell_0$ or $\ell_1$ penalties on the coefficients, as in \cite{aharonEtAl2006}
and \cite{lee06,mairalEtAl2010}, respectively.

In \cite{ranzato06,ranzato07aistat,ranzato07}, the authors proposed to learn a 
pair of encoding and decoding transformations for efficient
representation of natural images.
In this case the encodings of the training data are dense,
and sparsity is introduced by a further non-linear transformation between the
encoding and decoding modules.
Building on the idea of directly learning an encoding transformation,
we formulated the problem in the framework of regularized
optimization, by defining an $\ell_1$ penalized 
cost functional as in \cite{lee06,mairalEtAl2010}, augmented
by a {\em coding} term. 
This term penalizes the discrepancy between the optimal sparse representations
and those obtained by the inner products of the dual matrix and the 
input data.
The advantage of this approach in terms of performance stands out at 
evaluation time, when coding a new vector only requires a single matrix-vector
product.

The minimization of the proposed functional may be achieved by
block coordinate  descent, and we rely on proximal methods to
perform the three resulting inner optimization problems. Indeed, in
recent years different authors provided both theoretical and empirical
evidence that proximal methods may be used to solve the optimization
problems underlying many algorithms for $\ell_1$-based regularization 
and structured sparsity. A considerable amount of work has been devoted
to this topic within the context of signal recovery and image
processing. An extensive list of references and an overview of several
approaches can be found in 
\cite{beckEtAl2009,beckerEtAl2009,combettesEtAl2005,Dau04,nesterov05}, 
and \cite{duchi09} in the specific context of machine learning.  
Proximal methods have been used in the context of dictionary
learning by \cite{jenatton10}, where the authors are more focused on the
combination of dictionary learning with the notion of structured
sparsity. In particular they introduce a proximal operator for a
tree-structured regularization that is particularly relevant when one 
is interested in hierarchical models. 

Up to our knowledge, the present work is the first attempt to cast the problem of 
jointly learning a dictionary and its dual in the framework of regularized
optimization.
We show experimentally that PADDLE can recover the expected dictionaries
and duals, and that codes based on the dual matrix yields state-of-the-art 
classification performance while being much less computational intensive.

\section{Proximal methods for learning dual dictionaries}

In this section we lay down the problem setting and give a brief
overview on proximal methods, and how they can be applied to the 
problem at hand.
The actual PADDLE algorithm, with more details on its implementation,
is described in Section \ref{sec:algorithm}.

Let \mbox{$\bX=[\bx_1,\ldots,\bx_N]\in\RR^{d\times N}$} be the matrix
whose columns are the training vectors.
Our goal is to learn a primal dictionary 
\mbox{$\bD=[\bd_1,\ldots,\bd_K]\in\RR^{d\times K}$} (the {\em decoding}
or {\em synthesis} operator), and its dual
\mbox{$\bC=[\bc_1,\ldots,\bc_K]^T\in\RR^{K\times d}$} (the {\em encoding} 
or {\em analysis} operator), under some optimality conditions that we
will define in short. 
The columns of $\bD$ are the atoms of the dictionary, while the rows
of $\bC$ can be seen as filters that are convolved with an input
signal $\bx$ to encode it to a vector $\bu\in\RR^K$.
Both the atoms and the filters are constrained to have bounded norm to avoid
a trivial solution to the problem. 

Let now \mbox{$\bU=[\bu_1,\ldots,\bu_N]\in\RR^{K\times N}$} be the matrix 
whose columns are the encodings of the training data.
We set out to learn both $\bD$ and its dual $\bC$ by minimizing
\begin{eqnarray}
E(\bD,\bC,\bU)&=&\frac{1}{d}\|\bX-\bD\bU\|^2_F+\frac{\eta}{K}\|\bU-\bC\bX\|^2_F+\frac{2\tau}{K}\sum_{i=1}^N\|\bu_i\|_1,\label{eq:functional}\\
&s.t.&\|\bd_i\|^2,\|\bc_i\|^2\le 1\nonumber
\end{eqnarray}
where $\tau>0$ is a regularization parameter inducing sparsity in $\bU$,
while $\eta\ge 0$ weights the coding error with respect to the
reconstruction error.

Since the functional~\ref{eq:functional} is separately convex in each variable, we proceed by block coordinate descent (also
known as block nonlinear Gauss-Seidel method) \cite{luenberger2003},
iteratively minimizing first with respect to the encoding variables $\bU$ ({\em sparse coding}
step), and then to the dictionary $\bD$ and its dual $\bC$ ({\em dictionary update} step).
Such approach has been proved empirically successful~\cite{lee06}, and
its convergence towards a critical point of $E$ is guaranteed by
Corollary 2 of \cite{grippoEtAl2000}. 

The minimization steps both with respect to $\bU$ and w.r.t. $\bD$ and
$\bC$ cannot be solved explicitly, and therefore we are forced to find
approximate solutions using an iterative algorithm. In order to do so,
we strongly rely on the common structure of the three minimization
problems, consisting in the sum of a convex and differentiable function
and a non-smooth convex penalty term or constraint. 
The presence  of a non differentiable term makes the solution of the 
minimization problems non trivial. Proximal methods proceed by splitting
the contribution due to the non-differentiable part, giving  easily
implementable algorithms.  

In summary, a proximal (or forward-backward splitting) algorithm minimizes a function of type 
$E(\xi)=F(\xi)+J(\xi)$, where $F$ is convex and 
differentiable, with Lipschitz continuous gradient, while $J$ is lower
semicontinuous, convex and coercive. These assumptions on $F$
and $J$, required to ensure the existence of a solution, are fairly
standard in the optimization literature (see
e.g. \cite{combettesEtAl2005}) and are always satisfied in the setting
of dictionary learning for visual feature extraction. 
The non-smooth term $J$ is involved via its proximity operator, which can
be seen as a generalized version of a projection operator: 

\begin{equation}\label{eq:def_prox}
P(x)=\argmin_y\{J(y)+\frac 1 2 \|x-y\|^2\}.
\end{equation}
The proximal algorithm is given by combining the projection step with a
forward gradient descent step, as follows 
\begin{equation}
\xi^{p}=P\left(\xi^{p-1}-\frac{1}{2\sigma}\nabla F(\xi^{p-1})\right),
\label{eq:proximal}
\end{equation}

The step-size of the inner gradient descent is governed by the coefficient
$\sigma$, which can be fixed or adaptive, and whose choice will be 
discussed in Section \ref{sec:stepsize}. In particular, it can be shown that $E(\xi^p)$ converges 
to the minimum of $E$ if $\sigma$ is chosen
appropriately~\cite{combettesEtAl2005}. The convergence of $\xi^p$
towards a minimizer of $E$ is discussed below. 

\subsection{Sparse coding}

With fixed $\bD$ and $\bC$, we can apply algorithm \eqref{eq:proximal} to minimize the
functional \mbox{$E(\bU)=F(\bU)+J(\bU)$}, where
\begin{equation}
  F(\bU)=\frac{1}{d}\|\bX-\bD\bU\|^2_F+\frac{\eta}{K}\|\bU-\bC\bX\|^2_F~~\text{and}~~J(\bU)=\frac{2\tau}{K}\sum_{i=1}^N\|\bu_i\|_1.
\end{equation}

The gradient of the (strictly convex) differentiable term $F$ is
\[\nabla_{\bU}F=-\frac{2}{d}\bD^T(\bX-\bD\bU)+\frac{2\eta}{K}(\bU-\bC\bX),\]
while the proximity operator corresponding to  $J$ is the well-known soft-thresholding 
operator $\bS_{\lambda}$ defined component-wise as
\[(\bS_{\lambda}[\bU])_{ij}=\text{sign}(U_{ij})\max\{|U_{ij}|-\lambda,0\}.\]

Plugging the gradient and the proximal operator into the general equation 
\eqref{eq:proximal}, we obtain the following update rule:
\begin{equation}
  \label{eq:solveU}
  \bU^{p}=\bS_{\tau/K\sigma_U}\left[\left(1-\frac{\eta}{K\sigma_U}\right)\bU^{p-1}+\frac{1}{\sigma_U}\left(\frac{1}{d}\bD^T(\bX-\bD\bU^{p-1})+\frac{\eta}{K}\bC\bX\right)\right]
\end{equation}

\subsection{Dictionary update}

When $\bU$ is fixed, the optimization problems with respect to $\bD$ and $\bC$ are decoupled and can be solved separately. 

The quadratic constraints on the columns of $\bD$ and the rows of $\bC$ are 
equivalent to an indicator function $J$. 
Denoting by $B$ the unit ball in $\mathbb{R}^d$, the constraint on $\bD$ (respectively $\bC$) is formalized with $J$ being the indicator function of the set of matrices whose columns (resp. rows) belong to $B$.
Due to the fact that $J$ is an indicator function, in both cases the proximity operator is a projection operator.
Denoting by \mbox{$\pi(\bd)=\bd/\max\{1,\|\bd\|\}$} the projection on the unit ball in $\mathbb{R}^d$, let $\pi_D$ be the operator applying $\pi$ to the columns of $\bD$ and $\pi_C$ the operator applying $\pi$ to the rows of $\bC$.

Plugging the appropriate gradients and projection operators into Eq. 
\eqref{eq:proximal} leads to the update steps
\begin{eqnarray}
  \label{eq:solveD}
  \bD^p & = & \pi_D(\bD^{p-1}+\frac{1}{d\sigma_D}(\bX-\bD^{p-1}\bU)\bU^T),\\
  \label{eq:solveC}
  \bC^p & = & \pi_C(\bC^{p-1}+\frac{1}{K\sigma_C}(\bU-\bC^{p-1}\bX)\bX^T).
\end{eqnarray}


\subsection{Gradient descent step}
\label{sec:stepsize}

The choice of the step-sizes $\sigma_U$, $\sigma_D$ and $\sigma_C$ is
crucial in achieving fast convergence. 
Convergence of each minimization step is discussed in two senses: with
respect to the functional values, e.g. of
\mbox{$E(\bD^p)-\min_{\bD}E(\bD)$}, 
and with respect to the minimizing sequences, e.g. of $\bD^p$.

\subsubsection*{Fixed step-size}
In general, for $E=F+J$, one can choose the step-size to be equal, for all iterations, to the Lipschitz constant of $\nabla F$.
For $\nabla_{\bD}F$ and $\nabla_{\bC}F$ these constants can be evaluated explicitly, leading to \mbox{$\sigma_D=2\|\bU\bU^T\|_F/d$} and 
\mbox{$\sigma_C=2\|\bX\bX^T\|_F/K$}.
Such choices ensure linear rates of convergence in the values of $E$ 
\cite{beckEtAl2009} and convergence of the sequences $\bD^p$ and $\bC^p$ 
towards a minimizer \cite{combettesEtAl2005}, although no convergence rate 
is available.

A similar derivation would lead to $\sigma_U=2\|\frac 1 d \bD^T\bD+\frac \eta K \bI\|$, but in this case the positive definiteness of the matrix allows us to choose a step-size with faster convergence~\cite{rosasco09}.
Denoting by $a$ and $b$ be the minimum and maximum eigenvalues of $\bD^T\bD$,
choosing 
\[\sigma_U=\frac{a+b}{2d}+\frac{\eta}{K},\]
will lead to linear convergence rates in the value, as well as linear convergence for the sequence $\bU^p$ to the unique minimizer of $E(\bU)$~\cite{rosasco09}.

\subsubsection*{FISTA}

Improved convergence rates can be achieved in two ways: either through adaptive step-size choices (Barzilai-Borwein method), or by slightly modifying the proximal step as in FISTA \cite{beckEtAl2009}. 
The PADDLE algorithm makes use of the latter approach.

The FISTA update rule is based on evaluating the proximity operator with a weighted sum of the previous two iterates. 
More precisely, defining $a_1=1$ and $\phi^1=\xi^1$, the proximal step \eqref{eq:proximal} is replaced by
\begin{eqnarray}\label{eq:FISTA}
\xi^p&=&P\left(\phi^{p}-\frac{1}{2\sigma}\nabla F(\phi^{p})\right),\label{eq:FISTA1}\\
a_{p+1}&=&(1+\sqrt{1+4a_p^2})/2\label{eq:FISTA2}\\
\phi^{p+1}&=&\xi^p+\frac{a_p-1}{a_{p+1}}(\xi^p-\xi^{p-1}).\label{eq:FISTA3}
\end{eqnarray}

Choosing $\sigma$ as in the fixed step-size case, this simple modification allows to achieve quadratic convergence rate with respect to the values \cite{beckEtAl2009},
which is known to be the optimal convergence rate achievable using a first order method \cite{nemirovsky83}.
Although convergence of the sequences $\bD^p,\bC^p$ and $\bU^p$ is not
proved theoretically, there is empirical evidence that it holds in
practice~\cite{beckerEtAl2009}. Our experiments confirm this
observation.

\section{The PADDLE algorithm}
\label{sec:algorithm}

\begin{algorithm}[t]
\begin{algorithmic}[1]
    \REQUIRE $\bX\in\RR^{d\times N}$ \COMMENT{input data}
    \REQUIRE $\bU^0\in\RR^{K\times N}$, $\bD^0\in\RR^{d\times K}$, $\bC^0\in\RR^{K\times d}$ \COMMENT{initialization}
    \REQUIRE $rtol,\tau,\eta>0$, $T_{max}\in\ZZ^{+}$, $1\le H<T_{max}$ \COMMENT{algorithm parameters}
    \STATE $E^0=E(\bD^0,\bC^0,\bU^0;\bX,\eta,\tau)$
    \FOR{$t=1$ \TO $T_{max}$}
    \STATE $\bU^{t}=\argmin_{\bU}E(\bD^{t-1},\bC^{t-1},\bU;\bX,\eta,\tau)$, see Eq. \eqref{eq:solveU} \label{algo:solveU}
    \STATE possibly replace under-used atoms, see Sec. \ref{sec:algorithm} \label{algo:replace}
    \STATE $\bD^{t}=\argmin_{\bD}E(\bD,\bC^{t-1},\bU^t;\bX,\eta,\tau)$, see Eq. \eqref{eq:solveD}
    \STATE $\bC^{t}=\argmin_{\bC}E(\bD^t,\bC,\bU^t;\bX,\eta,\tau)$, see Eq. \eqref{eq:solveC}
    \STATE $E^t=E(\bD^t,\bC^t,\bU^t;\bX,\eta,\tau)$
    \STATE $E_H=\sum_{i=max\{p-H,0\}}^{t-1} E^i/H$
    \IF{$|E^t-E_H|/E_H < rtol$}
    \STATE break
    \ENDIF
    \ENDFOR
    \RETURN $\bU^t$, $\bD^t$, $\bC^t$
\end{algorithmic}
  \caption{PADDLE}
  \label{algo}
  \end{algorithm}

The complete algorithm we propose is outlined in Algorithm \ref{algo}.
As previously explained, the algorithm alternates between optimizing with
respect to $\bU$, $\bD$ and $\bC$.
These three optimizations are carried out employing the iterative projections
defined in equations \eqref{eq:solveU}, \eqref{eq:solveD} and 
\eqref{eq:solveC}, respectively, adapted according to
equations~(\ref{eq:FISTA1}--\ref{eq:FISTA3}). After the first iteration
of the algorithm, the three inner optimizations are initialized with the
results obtained at the previous iteration. This can be seen as an
instance of the popular {\em warm-restart} strategy.

During the iterations it may happen that, after the optimization with
respect to $\bU$, some atoms of $\bD$ are used only for few reconstructions, 
or not at all.
As suggested in \cite{aharonEtAl2006}, if the $i$-th atom $\bd_i$ is
under-used, meaning that only few elements of the $i$-th row of $\bU$ are
non-zero, we can replace it with an example that is not reconstructed
well (see line \ref{algo:replace}).
If $\bx_j$ is such an example, this can be achieved by simply setting 
$\bu_j=\be_i$, since at the next step $\bD$ and $\bC$ are estimated from 
$\bU$ and $\bX$.
In our experiments we only replaced atoms when not used at all.

The iterative procedure is stopped either upon reaching the maximum number
of iterations, or when the energy decreases only slightly with respect to
the last $H$ iterations. In our experiments we found out that, in
practice, after a few hundreds of outer iterations the convergence was
always reached. Indeed, in many cases only a few tens of iterations were
required. 

It is worth noting here, that in our implementation the algorithm 
optimizes with respect to all codes $\bu_i$ simultaneously, see line
\ref{algo:solveU}. 
Although not strictly necessary, it is a possibility we opted for, since
we confident it could prove advantageous in future hardware-accelerated
implementations. 
However, the algorithm can be easily implemented with a sequential 
optimization of the $\bu_i$.

A reference implementation in Python, together with scripts for
replicating the experiments of the following sections, are available
online at \url{http://slipguru.disi.unige.it/Research/PADDLE}.

\section{Experiments}

The natural application of PADDLE is in the context of learning
discriminative features for image analysis. Therefore, in the following
we report the experiments on standard datasets of digits and natural 
images, in order to perform qualitative and quantitative assessments 
of the recovered dictionaries for various choices of the parameters.
Furthermore, we discuss the impact of the feature vectors obtained from
a learned $\bC$ on the accuracy of an image classifier. \\
Preliminarily, we also present a set of experiments on synthetic data,
aimed at understanding how well the algorithm can recover a
dictionary and its dual in a more controlled setting.

\subsection{Synthetic data}

\begin{figure}[!t]
\begin{center}
\begin{tabular}{cc}
  \includegraphics[height=.3\textheight]{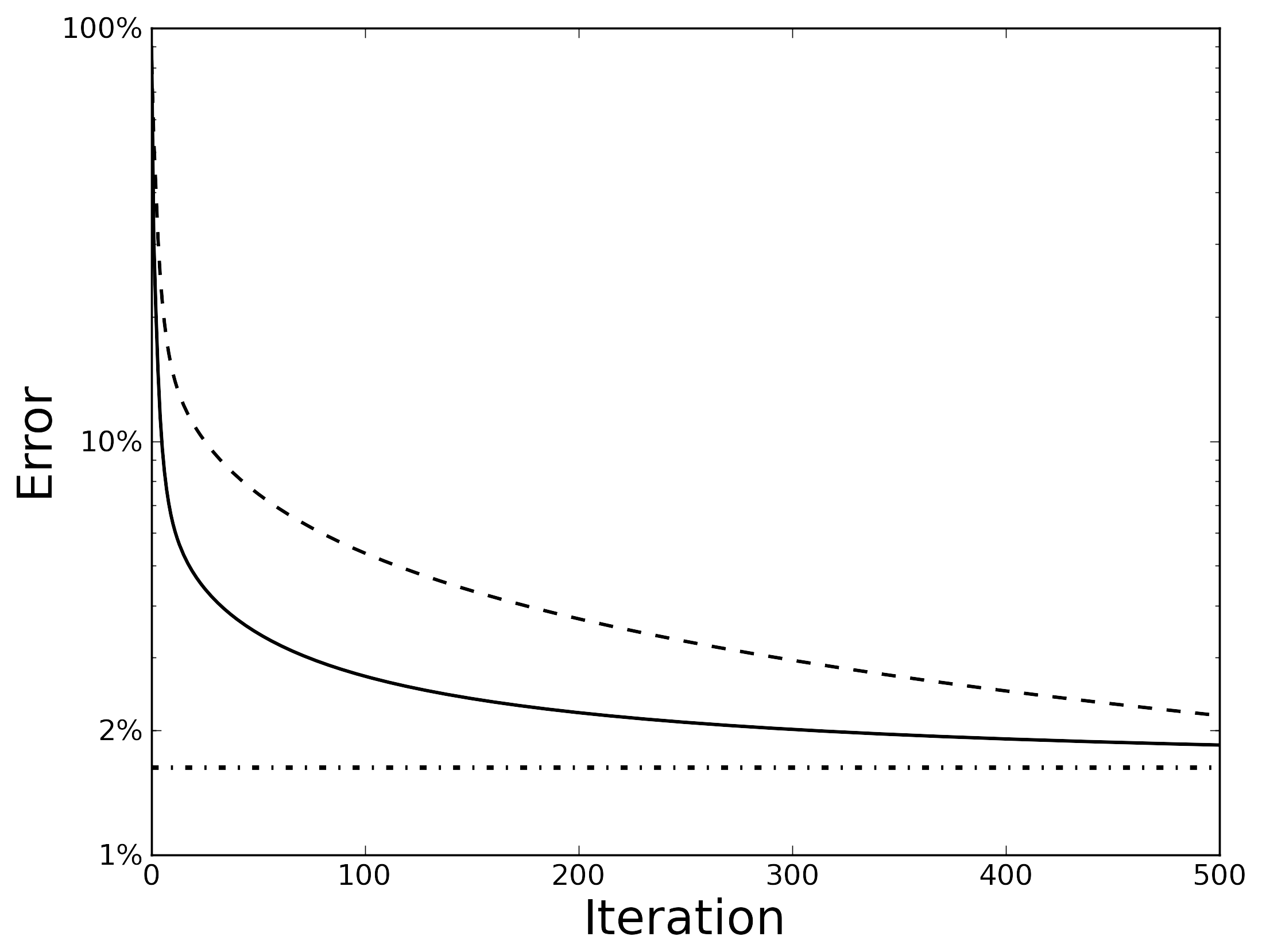} &
  \includegraphics[height=.3\textheight]{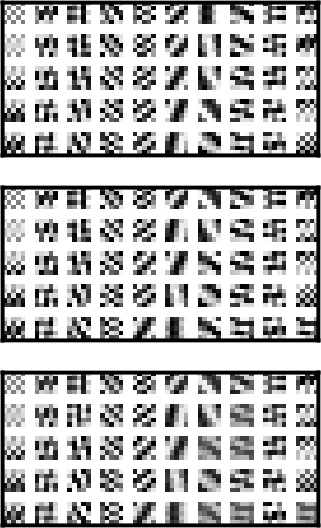} 
\end{tabular}
\end{center}
\caption{Results of the synthetic experiments. ({\em left}) Experiment for
 $K<d$ and $\tau=0$. The reconstruction error (solid line) converges to
 the optimal error (dash-dotted line) and the distance between $\bC$ and
 $\bD^T$  (dashed black) decades with it.
  ({\em right}) Synthetic experiments with a tight frame. From top to
 bottom, we show the original dictionary, the recovered dictionary $\bD$
 and recovered dual $\bC^T$. The order in not necessarily the same
  as in the original, and some of the atoms could be inverted.
}
\label{fig:synth}
\end{figure}

The first synthetic experiment has been performed with $K<d=25$ by generating
$N=10^4$ training vectors as linear combinations of 15 random vectors.
The data have been corrupted by additive Gaussian noise.
We have run the algorithm  with $K=15$, $\eta=1$ and a small $\ell_2$ regularization 
term on the coefficients $\bU$, in order to ensure stability.
In this setting we expected the algorithm to recover a dictionary close
to the basis of the first 15 principal components, as well as converging
to a dual $\bC$ close to the transpose $\bD^T$.
In the experiments the distance between $\bD$ and the PCA basis, assessed by 
the largest principal angle between the spanned subspaces, has decreased
under $10^{-7}$ after very few iterations.
In Figure \ref{fig:synth}, left image, we also show that the reconstruction 
error converges to the minimum achievable with the first $K$ principal 
components and that the dual $\bC$ also converges to $\bD^T$.
The distance between $\bD$ and its dual has been computed as the mean value of 
\mbox{$1-|\bd_i\cdot\bc_i|/(\|\bd_i\|\|\bc_i\|)$} (where the $\bc_i$ are 
the rows of $\bC$).


In the second experiment we have constructed a tight frame, shown
in Figure \ref{fig:synth} (top, right), and generated $N=10^4$ training
vectors as random superposition of 3 frame elements.
The data have been again corrupted with additive Gaussian noise, and the
algorithm run with $K=50$ (the same number of frame elements), 
$\eta=1$ and $\tau=0.5$.
The reconstruction error has reached immediately the minimum achievable with
the true generating frame, and the distance between the $\bC$ and
$\bD^T$, computed as in the previous experiment, has got smaller than $10^{-2}$.
In Figure \ref{fig:synth} (right) we show the generating frame and
compare it with the recovered dictionary $\bD$ and dual $\bC^T$.
As apparent, most elements are present in both dictionaries (possibly 
inverted).


\subsection{Benchmark datasets: digits and natural images}
\label{sec:benchmark}

In the next experiments we have applied PADDLE to two sets of images
widely used as benchmarks in computer vision: the Berkeley segmentation
dataset \cite{bsd} and the MNIST dataset \cite{mnist}. The aim of this
section is to offer a qualitative and quantitative assessment of the
dictionaries recovered from these two classes of images, with realistic
sizes of the training set. 

\subsubsection*{Berkeley segmentation dataset}

Following the experiment in \cite[Sec.~4]{ranzato06}, we have extracted
a random sample of $10^5$ patches of size 12x12 from the Berkeley
segmentation data set \cite{bsd}.
The images intensities have been centered by their mean and normalized  
dividing by half the range (125). The patches have been separately
recentered too.

\begin{figure}[!t]
  \begin{center}
   \includegraphics[width=.32\textwidth]{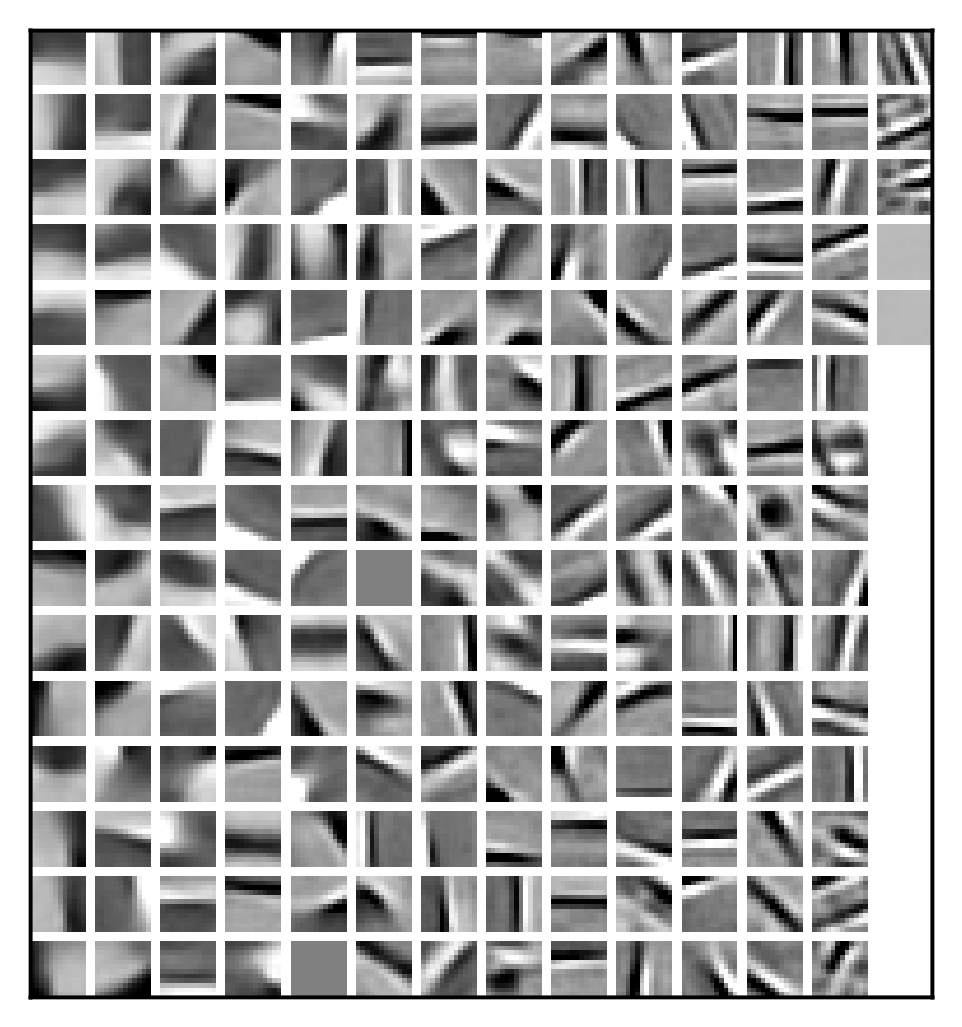}
   \includegraphics[width=.32\textwidth]{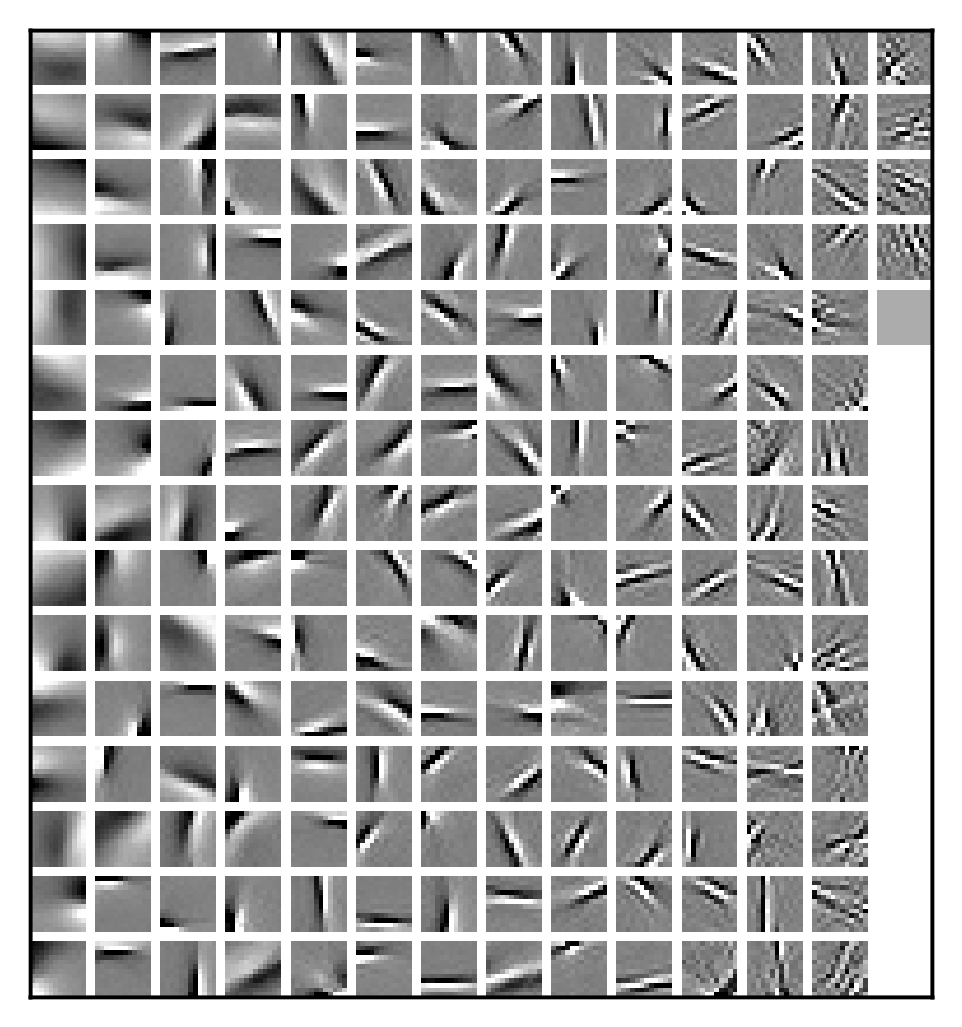}
   \includegraphics[width=.32\textwidth]{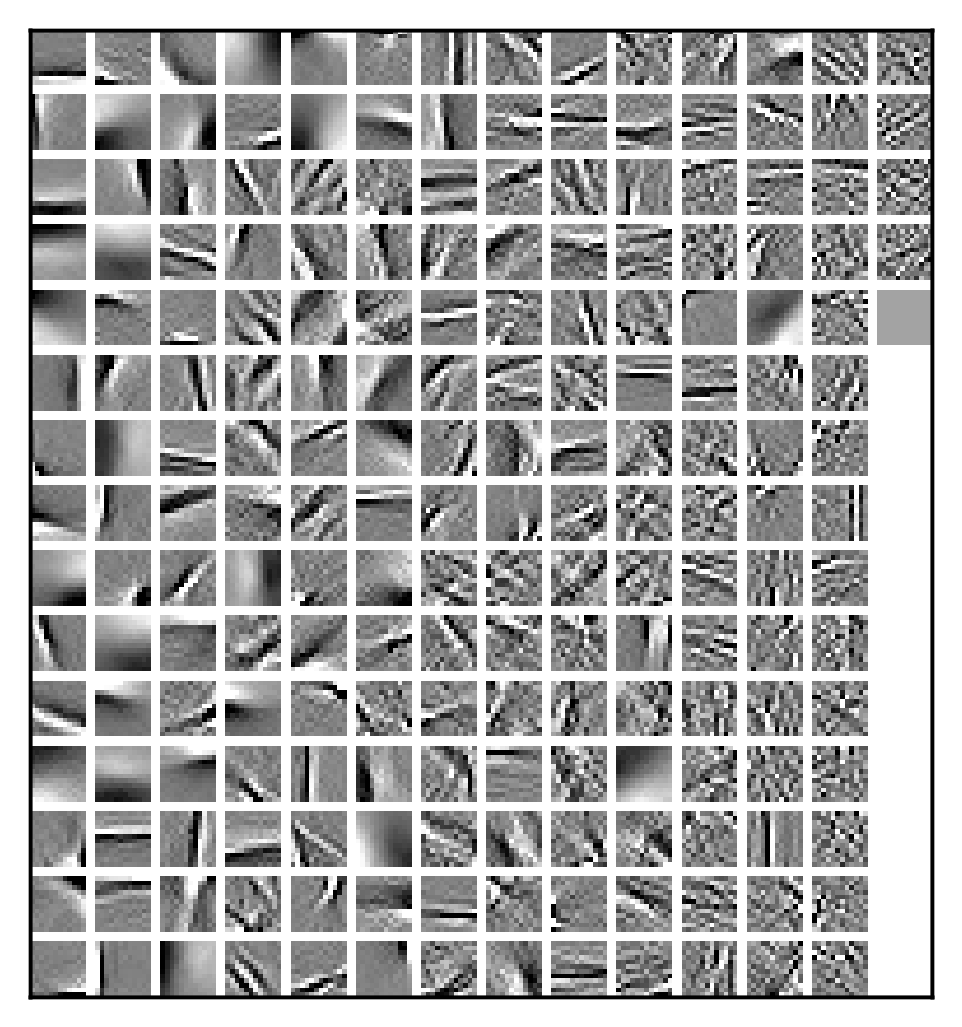}
   \\
   \includegraphics[width=.32\textwidth]{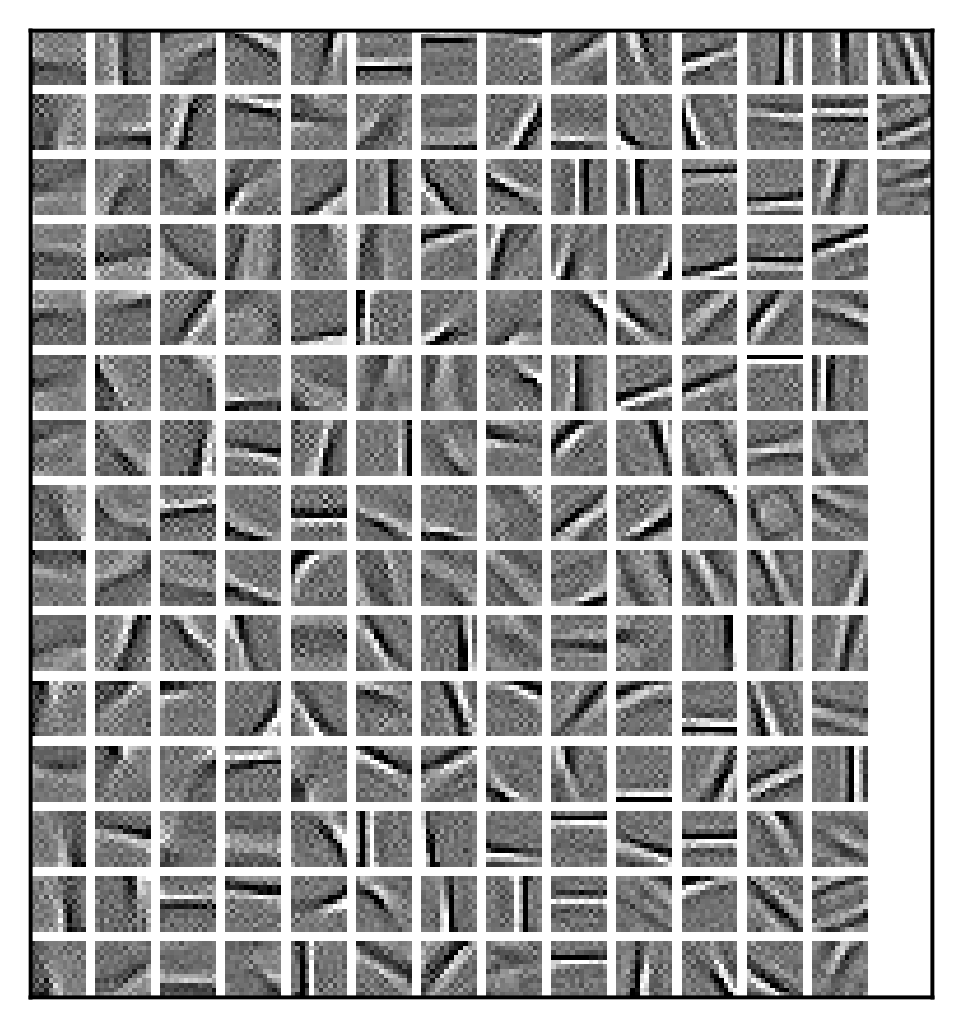}
   \includegraphics[width=.32\textwidth]{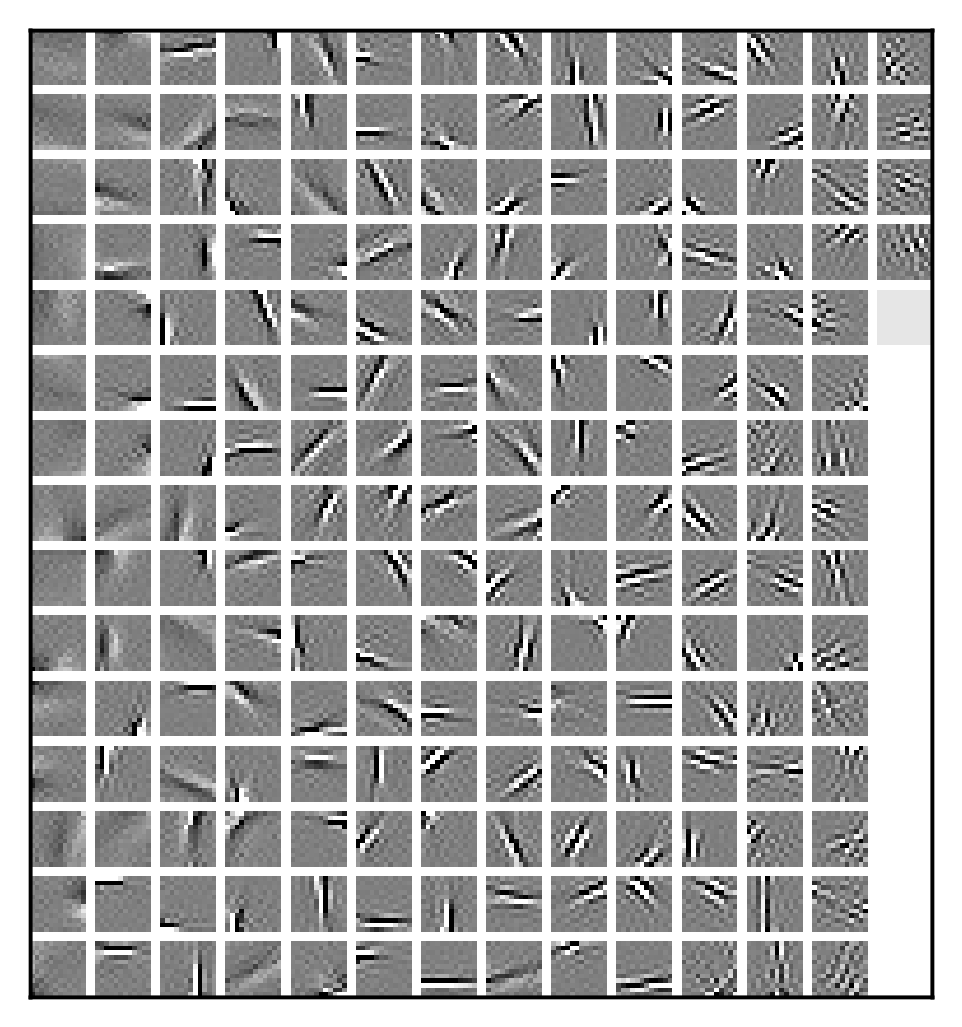}
   \includegraphics[width=.32\textwidth]{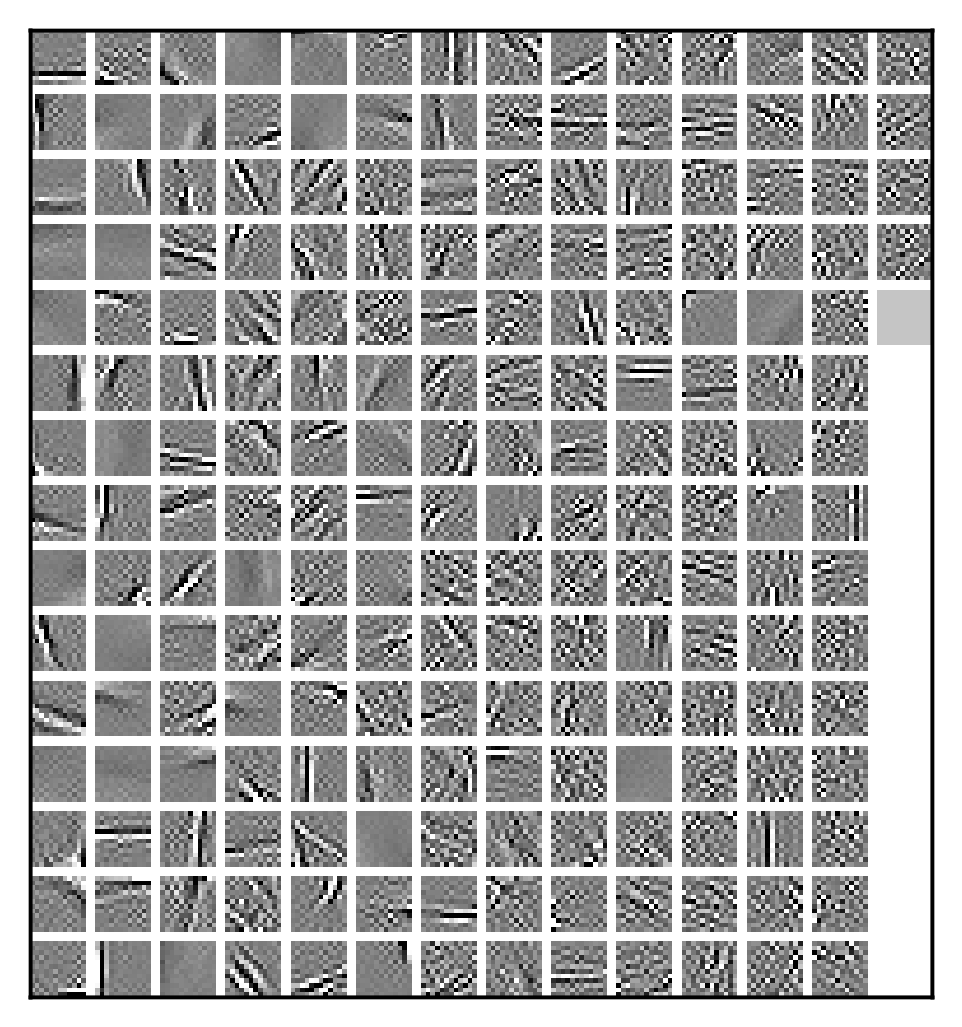}
   \caption{Decoding (top) and coding (bottom) matrices $\bD$ and $\bC$ 
   learned from $100,000$ patches of size $12\times 12$ randomly sampled from 
   the Berkley segmentation dataset.
   The dictionaries have been learned with different values of $\tau$:
   from left to right, $1$, $0.1$ and $0.02$.
   The atoms are in column-major order from the most used to the least one.
   The codes corresponding to the three dictionaries pairs uses on average 
   6.6, 65.9 and 139.8 atoms, respectively.}
   \label{fig:bsd_dict}
  \end{center}
 \end{figure}

We have run PADDLE over a range of values for $\tau$, and both with 
coding error ($\eta=1$) and without ($\eta=0$).
The relative tolerance for stopping has been set to $10^{-4}$.
The reconstruction error achieved at the various level of sparsity ($\tau$), 
both with and without the coding error, have been constantly lower than the 
reconstruction error achievable with a comparable number of principal
components. \\
In Figure \ref{fig:bsd_dict} we show images of the recovered dictionary,
together with their duals. An interesting effect we report here is that
different levels of sparsity in the coding coefficients also affects the
visual patterns of the dictionary atoms. The more sparse is the
representation the more the (very few) atoms tends to look like simple
partial derivatives of a $2D$ Gaussian kernel, i.e. the dictionary tends
to adapt poorly to the specific set of data. On the contrary, with a
less sparse representation, a larger number of the atoms seem to encode
for more structured local patterns or specific textures present in the
dataset. 


\subsubsection*{MNIST dataset}

\begin{figure}[!t]
 \begin{center}
  \begin{tabular}{cccc}
   {\bf Initialization} &   {\bf After $1$ iteration} &    {\bf After $4$ iterations} & {\bf At convergence} \\
   \includegraphics[width=.2\textwidth]{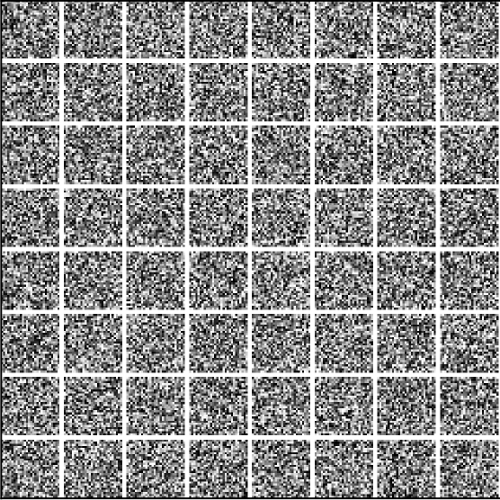} &
       \includegraphics[width=.2\textwidth]{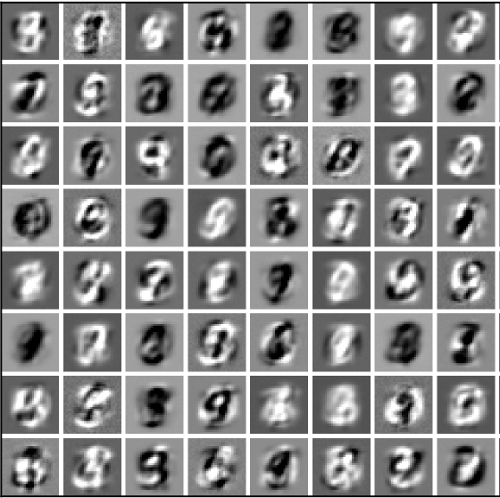} &
   \includegraphics[width=.2\textwidth]{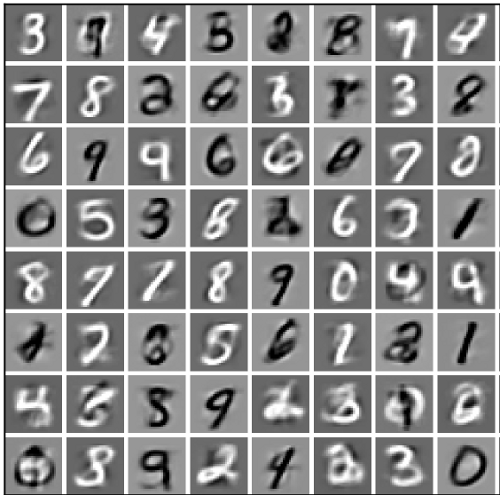} & \includegraphics[width=.2\textwidth]{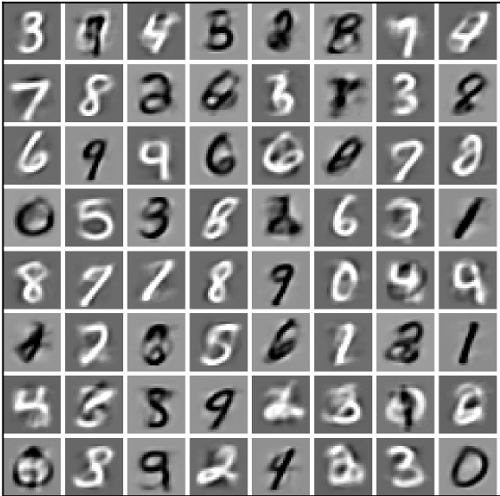}
  \end{tabular}  \\
  \vspace{.1cm}
  \includegraphics[width=.6\textwidth]{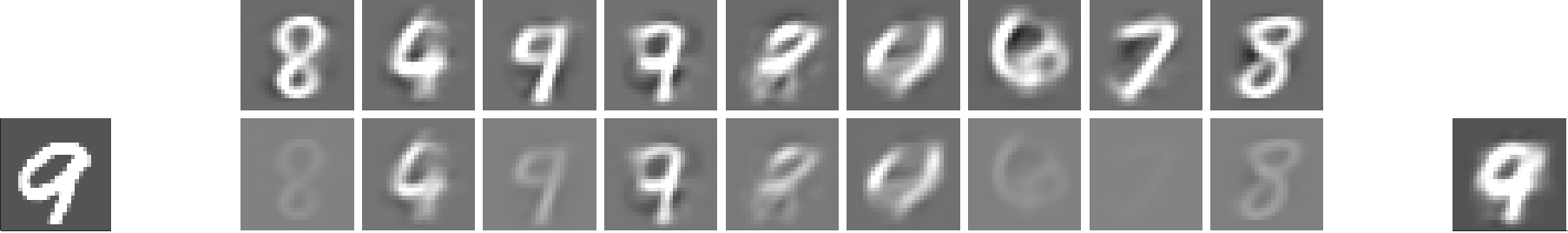}
  \caption{Experiments on MNIST dataset. See section \ref{sec:benchmark} for 
    details.}\label{fig:mnist} 
 \end{center}
\end{figure}

Next we have tested the algorithm on the $50,000$ training images of the
popular MNIST data set \cite{mnist}, which is a collection of 
$28\times 28$ quasi binary images of handwritten digits. According to
the experiments described in~\cite{ranzato06}, we have trained the
dictionary with $200$ atoms which is very likely to represent an
overcomplete setting because the true dimensionality of the images is
definitely less than $784$. 
All the images have been pre-processed by mapping their range into the
interval $[0,1]$.  
The results obtained are consistent with those already reported in the
literature. In particular, the learned dictionary $\bD$ comprises the most
representative digits from which it is possible to reconstruct all the
others with a low approximation error. In Figure~\ref{fig:mnist}~(bottom) we
show how the exemplar digit on the left can be expressed in terms of the
small subset  of the atoms in the middle, obtaining the approximate image on the
right. As expected, the actual number of non-zero coefficients is
extremely low if compared with the size of the dictionary. In the two
rows in the middle, we first report all the dictionary atoms with
non-zero coefficients and then we weight them with respect to their
relevance in the reconstruction.\\
A second, more interesting aspect is the fast empirical convergence of
the algorithm with such a well-structured dataset, as shown in
Figure~\ref{fig:mnist} (on the top). 
The initial dictionary has been built with random patches. After the first
iteration it was already possible to inspect some digits, and the amount
of change decreased rapidly reaching a substantial convergence after
only a few iterations. The dictionary after $20$ iterations (corresponding
to the full convergence) was almost identical to the one after $4$
iterations only. 

\subsection{Classification}

In this last group of experiments we have focused on the impact of using
the dictionaries learned with the PADDLE algorithm in a classification
context. More specifically, we have investigated the discriminative power
of the sparse coding associated to the dictionary $\bD$ and its dual
$\bC$ when used to represent the visual content of an image. The
goal of the experiments has been to build a classifier to assign each
image to a specific semantic class. In practice, we replicated the 
experimental setting of~\cite{yang09}. The classification results,
reported in Table~\ref{tab:classification}, show that 
the performance we have obtained using a representation computed with PADDLE
is essentially the same -- i.e. the results are not distinguishable within
one standard deviation -- as the one obtained with learned dictionary 
used by the authors in the original paper. 

\begin{table}[!th]
 \begin{center}
  \begin{tabular}{|r|c|c|c|}
   \hline 
   {\bf Encodings} & $\bD$ & $\bC$ & \cite{yang09} \\
   \hline
   {\it Mean accuracy} &  0.987 & 0.984  & 0.985 \\
   {\it Standard deviation} & 0.008 & 0.008  & 0.008 \\
   \hline
  \end{tabular}
  \caption{Classification performances obtained on a subset of the
  popular Caltech101 dataset comprising two classes.}
  \label{tab:classification} 
 \end{center}
\end{table}

The results obtained with the dictionary $\bC$ are especially
encouraging if one consider the substantial gain in the computational time 
required
to compute the sparse codes, with a fixed dictionary, for each new
input image. In our experiments we have been able to process each image
in less than $0.21$ seconds, while it took $2.3$ seconds per image, on
the same machine, to use the implementation of the feature-sign search
algorithm \cite{lee06} provided with the software package ScSPM
(available at
\texttt{http://www.ifp.illinois.edu/~jyang29/ScSPM.htm}). 
Indeed, regardless the specific implementation of the sparse
optimization method, it is easy to see that using $\bC$ is always the
best choice since it requires just one matrix-vector multiplication.

\section{Conclusion}

We have proposed a novel algorithm based on proximal methods to learn
a dictionary and its dual, that can be used to compute sparse overcomplete
representations of data.
The experiments have shown that for image data the algorithm yields 
representations with good discriminative power.
In particular, the dual dictionary can be used to efficiently compute the 
representations by means of a simple matrix-vector multiplication, without 
any loss of classification accuracy.
We believe that our method is a valid contribution towards building 
robust and expressive dictionaries of visual features.




\small {\newpage
\bibliographystyle{unsrt}

}

\end{document}